\PassOptionsToPackage{table}{xcolor}
\documentclass{article}

\usepackage{iclr2027_conference,times}
\iclrfinalcopy

\usepackage{amsmath,amsfonts,bm}

\def\eqref#1{equation~\ref{#1}}

\def\1{\bm{1}}

\DeclareMathAlphabet{\mathsfit}{\encodingdefault}{\sfdefault}{m}{sl}
\SetMathAlphabet{\mathsfit}{bold}{\encodingdefault}{\sfdefault}{bx}{n}

\usepackage{enumitem}
\usepackage{latexsym}
\usepackage{xspace}
\usepackage{amssymb}
\usepackage{float}
\usepackage{wrapfig}
\usepackage[T1]{fontenc}
\usepackage{amsmath}
\usepackage{booktabs}      
\usepackage{graphicx}      
\usepackage[utf8]{inputenc}
\usepackage{hyperref}
\usepackage{url}

\usepackage{microtype}
\usepackage{amsthm}
\newtheorem{proposition}{Proposition}

\newcommand{\method}{\textsc{CodeBlock}\xspace}

\floatstyle{plaintop}
\newfloat{algorithm}{tbp}{loa}
\floatname{algorithm}{Algorithm}
\newcounter{ALGline}
\newenvironment{algorithmic}[1][0]{%
  \setcounter{ALGline}{0}%
  \begin{list}{}{%
    \setlength{\leftmargin}{0pt}%
    \setlength{\rightmargin}{0pt}%
    \setlength{\itemsep}{0pt}%
    \setlength{\parsep}{0pt}%
    \setlength{\topsep}{0pt}}%
  \item\relax
}{\end{list}}
\newcommand{\ALGlineprefix}{%
  \refstepcounter{ALGline}%
  \makebox[2.1em][r]{\scriptsize\arabic{ALGline}:}\hspace{0.5em}}
\newcommand{\Require}{\par\noindent\textbf{Require:}\hspace{0.5em}\ignorespaces}
\newcommand{\Ensure}{\par\noindent\textbf{Ensure:}\hspace{0.5em}\ignorespaces}
\newcommand{\State}{\par\noindent\ALGlineprefix\ignorespaces}
\newcommand{\Comment}[1]{\hfill$\triangleright$~#1}
\newcommand{\For}[1]{\State\textbf{for} #1 \textbf{do}\par\begingroup\advance\leftskip by 1.25em\relax}
\newcommand{\EndFor}{\par\endgroup\State\textbf{end for}}
\newcommand{\If}[1]{\State\textbf{if} #1 \textbf{then}\par\begingroup\advance\leftskip by 1.25em\relax}
\newcommand{\ElsIf}[1]{\par\endgroup\State\textbf{else if} #1 \textbf{then}\par\begingroup\advance\leftskip by 1.25em\relax}
\newcommand{\EndIf}{\par\endgroup\State\textbf{end if}}
\newcommand{\Return}{\textbf{return}\hspace{0.5em}}

\title{\method: Learning to Supervise Code at the Right Granularity}

\author{
\textbf{Zhijie Deng}\textsuperscript{1}, \hspace{0.5mm}
\textbf{Ling Li}\textsuperscript{1}, \hspace{0.5mm}
\textbf{Jinlong Pang}\textsuperscript{3}, \hspace{0.5mm}
\textbf{Kaiqin Hu}\textsuperscript{4},
\hspace{0.5mm}
\textbf{Qi Xuan}\textsuperscript{5}, \hspace{0.5mm}
\textbf{Xuming Hu}\textsuperscript{1,2}, \\
\textbf{Zhaowei Zhu}\textsuperscript{5,6},
\hspace{0.5mm}
\textbf{Jiaheng Wei}\textsuperscript{1,$\dagger$}
\\
\textsuperscript{1}The Hong Kong University of Science and Technology (Guangzhou) \\
\textsuperscript{2}The Hong Kong University of Science and Technology \\
\textsuperscript{3}UC Santa Cruz \quad
\textsuperscript{4}Ant Group \quad
\textsuperscript{5}BAIA, ZJUT \quad
\textsuperscript{6}D5Data.ai \\
\texttt{\small zdeng190@connect.hkust-gz.edu.cn, jiahengwei@hkust-gz.edu.cn}
}

\begin{document}
\maketitle
\begin{abstract}
Supervised fine-tuning of code LLMs typically applies uniform cross-entropy loss to all response tokens, implicitly assuming that every token provides equally useful learning signals. Recent token-level selection methods challenge this assumption in natural-language SFT by supervising only high-value tokens. However, such pointwise selection can fragment the syntactic structures and program dependencies of code, leaving supervision scattered across incomplete code units. In experiments on 30K code instruction–response pairs, under the same 10\% token budget, supervising complete coding blocks improves average performance by 11.7 points over prior approaches that supervise isolated tokens. Motivated by this observation, we propose \method, a structure-aware sparse supervision framework that uses complete, parser-aligned coding items as the basic units of supervision. \method constructs coding items from high-quality code instruction data, estimates their supervision utility using GCE, which is more robust to low-probability tokens, and further adjusts their supervision priority using data-flow reach and bridge signals. During training, the full response is retained as context, while loss is applied only to the selected supervision units. Experiments show that partial supervision over only about 6.9\% of response tokens consistently outperforms full-token SFT across all five model settings, suggesting that effective code SFT depends not only on identifying high-value supervision, but also on allocating it at the appropriate structural granularity.
\end{abstract}

\section{Introduction}
Supervised fine-tuning (SFT) is a standard way to adapt code LLMs to instruction-following and program-generation tasks \citep{hui2024qwen2, guo2024deepseek}. As code instruction corpora grow, improving SFT increasingly depends on extracting dense, reliable, and high-marginal-utility supervision rather than simply adding more examples \citep{wei2024magicoder, li2024quantity, yu2025co}. This has motivated data selection methods that filter high-quality, less noisy, or more diverse instruction--response pairs before training \citep{chen2023alpagasus, xia2024less, deita}.
However, most existing methods remain sample-level: they keep or discard entire instruction--response pairs, while still assigning uniform cross-entropy supervision to every token in a selected response \citep{chen2025mig, wu2024rose}.

\begin{figure}[t]
    \vspace{-0.4in}
    \centering
    \begin{minipage}[t]{0.49\linewidth}
        \vspace{0pt}
        \centering
        \includegraphics[width=0.90\linewidth]{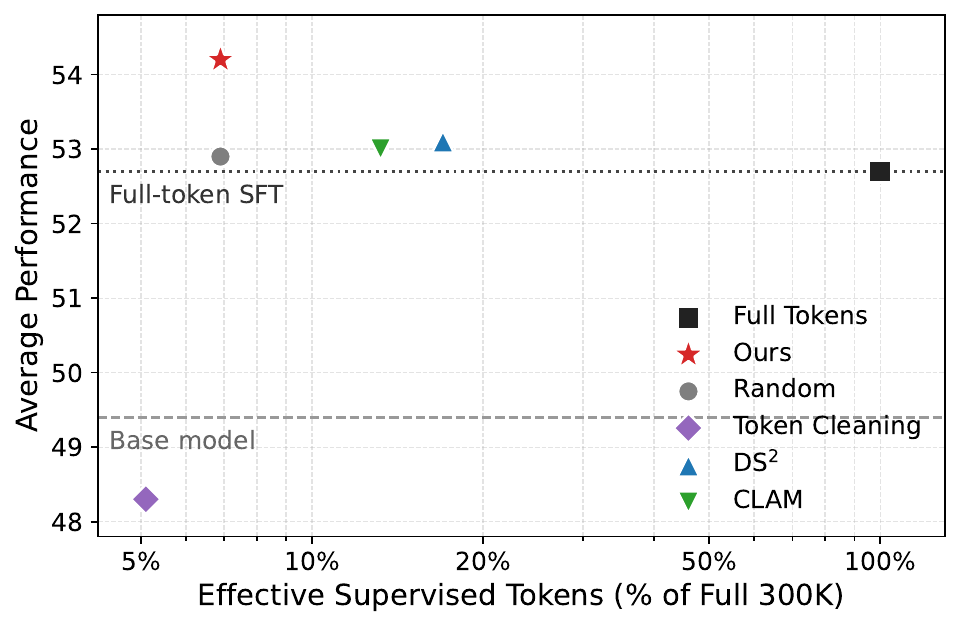}
        \vspace{-0.15cm}
        \caption{
        Performance--efficiency trade-off on Qwen2.5-Coder-1.5B-Instruct. \method achieves the highest average performance with only 6.9\% effective supervised tokens, showing a better trade-off than full-token SFT and sparse-selection baselines.
        }
        \label{fig:avg_token_trade_off}
    \end{minipage}
    \hfill
    \begin{minipage}[t]{0.49\linewidth}
        \vspace{0pt}
        \centering
        \vspace{0.45cm}
        \includegraphics[width=\linewidth]{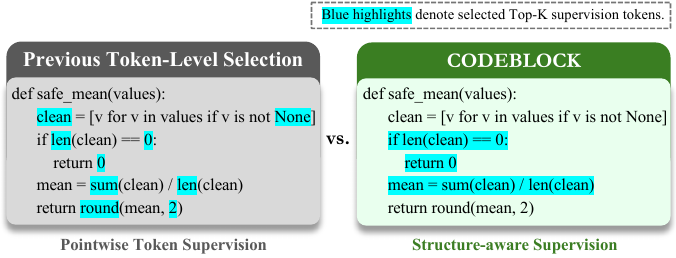}
        \vspace{0.45cm}
        \caption{
        Comparison between previous token-level selection and \method.
        While prior methods select isolated high-score tokens, \method supervises materialized parser-guided coding items and uses their data dependencies for selection.
        }
        \label{fig:intro}
    \end{minipage}

\end{figure}

This assumption has already been challenged in natural-language SFT. Recent token-level selection methods challenge the need for dense supervision by keeping only high-value tokens, often estimated by pointwise loss or excess-loss scores  \citep{pang2025tokencleaning, lin2024rho}. These methods are effective in natural-language settings because individual tokens can often be treated as approximate local learning units \citep{lin2024critical, qin2025sstoken, fu2025tshirt}. However, directly transferring pointwise token selection from natural-language settings to code SFT is unreliable. Unlike ordinary text, code is organized by a formal grammar, exposes mechanically recoverable local constructs and definition--use relations, and is evaluated by executing the generated program \citep{allamanis2017learning, guo2020graphcodebert}. An isolated variable name or operator does not specify the assignment, branch condition, API call, or return expression in which it participates. Figure~\ref{fig:intro} illustrates this granularity mismatch: pointwise token selection tends to pick scattered tokens from different statements, whereas structure-aware selection preserves materialized parser-guided items that form locally meaningful program evidence. In Sec.~\ref{sec:motivation}, we connect this structural distinction to the learning objective: the summed teacher-forced loss over a materialized item bounds the parser or test failure of a reference-conditioned local greedy replacement, and full-item supervision places every term of this bound inside the masked objective. A matched-budget diagnostic then verifies that this granularity is beneficial in downstream code generation. Together, these results motivate using token scores to locate valuable evidence while allocating the final supervision at coding-item granularity.

To this end, we propose \method, a structure-aware sparse supervision framework for code LLM fine-tuning. \method uses high-scoring tokens only as anchors for locating useful code evidence: it constructs parser-guided coding items, scores each item by the concentration of informative logic tokens using generalized cross-entropy, and then reranks items with lightweight data-flow signals that measure downstream influence and dependency-path connectivity. During training, the full response remains available as autoregressive context, while the loss is applied only to selected code items and informative natural-language tokens. As shown in Figure~\ref{fig:avg_token_trade_off}, across six code-generation benchmarks and five model settings, \method consistently matches or improves full-token SFT while using only about 6.9\% supervised response tokens, and achieves the best or second-best average performance among competitive selection baselines.

The contributions of this paper are as follows:

\begin{itemize}
\item On a 30K diagnostic subset at the top-10\% token budget, pointwise selection leaves 92.3\% of touched coding items only partially covered. Under the same code-token budget, supervising complete materialized items improves average performance by 11.7 points over isolated top-token supervision. A reference-conditioned local bound further characterizes the loss terms omitted by partial item supervision.

\item We propose \method, a structure-aware sparse supervision framework that selects coding items rather than individual tokens, combining item partitioning, GCE-based utility scoring, and data-flow-aware reranking.

\item Experiments across six code-generation benchmarks show that \method matches or outperforms full-token SFT with only about 6.9\% supervised response tokens, achieving a stronger performance--efficiency trade-off than competitive baselines.
\end{itemize}

\section{Related Work}
\textbf{LLM Data Selection.}
Instruction-tuning performance depends strongly on data quality and composition~\citep{deng2025lm}. Existing methods select training samples using LLM-based quality or difficulty signals, as in AlpaGasus, DEITA, IFD, DS2, D3 and Scar~\citep{chen2023alpagasus,deita,li2024quantity,ds2,zhang2025d3,li2025scar}; influence-based objectives, as in LESS and NICE~\citep{xia2024less,wang2025nice}; or diversity and information criteria, as in NovelSum and MIG~\citep{yang2025novelsum,chen2025mig}. Despite different selection signals, these methods operate primarily at the sample level and retain complete responses once an instruction--response pair is selected.

For code LLMs, data curation similarly emphasizes execution validity, redundancy reduction, and quality--diversity trade-offs. SelfCodeAlign retains self-generated responses through sandbox verification~\citep{wei2024selfcodealign}, while CSAM, distribution-consistent selection, and XCoder filter or select code instruction data based on redundancy, distribution and diversity, or instruction complexity and response quality~\citep{csam,lyu2025efficient,xcoder}. These approaches improve which instruction--response pairs enter training, but still treat each selected response as a uniform supervision unit. In contrast, \method studies supervision within selected code responses, determining which complete code fragments should contribute training gradients while retaining the full response as context.

\textbf{Fine-Grained Supervision Selection and Data Cleaning.}
Beyond sample-level filtering, recent works increasingly exploit token-level training signals. Rho-1 selectively applies loss to valuable pretraining tokens, while Token Cleaning and TOSS mask uninformative or unsafe SFT tokens~\citep{lin2024rho,pang2025tokencleaning,li2026tokensafe}. T-SHIRT incorporates token informativeness into hierarchical instruction-data selection, TokenTune jointly models token- and sample-level utility with gated optimization, and MSL masks invalid output tokens in recommendation fine-tuning~\citep{fu2025tshirt,lin2026tokentune,wang2025msl}. These works demonstrate the value of supervision below the sample level, but largely score or mask tokens as pointwise units. For code responses, supervision utility additionally depends on syntactic units, definition--use relations, and data-flow dependencies, motivating structure-aware selection over complete code fragments.

\section{Preliminary: Next-Token Prediction with Sparse Supervision}

Given an instruction-tuning dataset
$\mathcal{D}=\{(x_i,y_i)\}_{i=1}^N$,
where $p_\theta$ denotes the model's next-token distribution,
$x_i$ is the instruction, and
$y_i=\{y_{i,t}\}_{t=1}^{T_i}$ is the response,
we formulate supervised fine-tuning with a token-level supervision mask.
For each response token $y_{i,t}$, let $m_{i,t}\in\{0,1\}$ indicate whether
this token contributes to the training loss. The masked next-token prediction
objective is defined as:
{\small
\begin{equation}
\mathcal{L}_{\mathrm{mask}}(\theta)
=
-\frac{
\sum_{i=1}^{N}\sum_{t=1}^{T_i}
m_{i,t}\log p_{\theta}(y_{i,t}\mid x_i,y_{i,<t})
}{
\sum_{i=1}^{N}\sum_{t=1}^{T_i}m_{i,t}
}.
\end{equation}
}

This formulation generalizes standard full-token SFT and sparse supervision. 
When $m_{i,t}=1$ for all response tokens, the objective reduces to standard SFT, where every response token is supervised. 
Sparse supervision instead sets $m_{i,t}=0$ for most tokens by default and only enables selected tokens with $m_{i,t}=1$. 
Tokens with $m_{i,t}=0$ are still kept in the autoregressive context, but they are ignored in the loss computation.

Prior fine-grained selection methods~\citep{pang2025tokencleaning,lin2024rho} usually construct $m_{i,t}$ by scoring individual tokens, for example using token-level loss or excess-loss signals, and then retaining high-scoring tokens. 
However, this pointwise view is less suitable for code, where token roles are organized by parser-defined constructs and definition--use dependencies. 
This motivates our later use of materialized parser-guided coding items as the basic units for sparse supervision.

\section{Motivating Analysis}
\label{sec:motivation}
\begin{wrapfigure}[10]{r}{0.52\textwidth}
    \vspace{-3.8em}
    \centering
    \includegraphics[width=\linewidth]{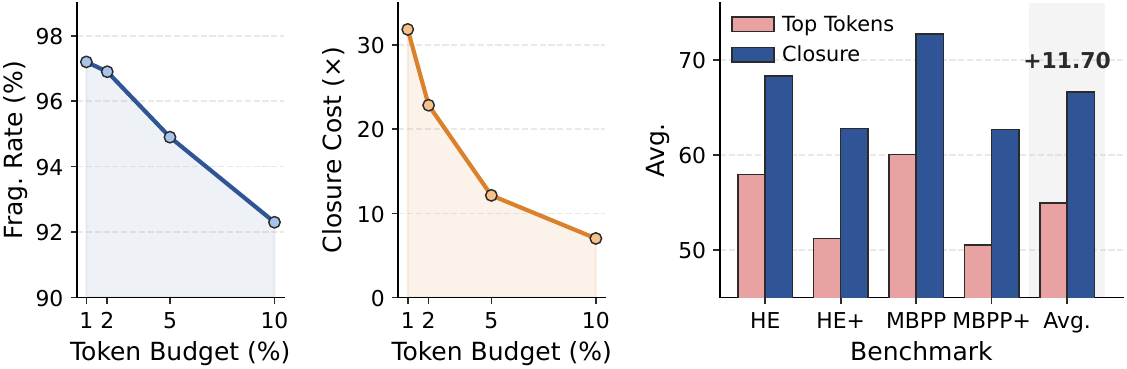}
    \caption{
Motivating analysis of token-level sparse supervision in code.
Left: fragmentation rate of coding items under different token budgets.
Middle: closure cost for expanding selected tokens into materialized local fragments.
Right: same-budget mini-SFT results comparing isolated top-token supervision with materialized-item supervision.
}
    \label{fig:motivating_analysis}
\end{wrapfigure}
Token-level selection assumes that high-scoring tokens are suitable training targets, but this assumption may not hold for code, where token semantics often depend on surrounding syntactic and
dependency structures. We therefore first examine whether high-scoring code tokens can directly serve as sparse supervision units.

\noindent \textbf{Definition of coding items.}
For each code response $y_i$, we partition its code regions into parser-guided local units $\mathcal{U}_i=\{u_{i,1},\ldots,u_{i,K_i}\}$. A coding item $u$ is a coherent local computation, such as an assignment, branch condition, API call, or return expression. We denote its core logic tokens (identifiers, literals, operators, and API names) by $C(u)$ and its contiguous materialized span by $M(u)$; the latter also retains associated boundary and protected syntax positions. Thus $C(u)\subseteq M(u)$, and an item is fully materialized when all positions in $M(u)$ are supervised. Section~\ref{sec:gce_scoring} gives the operational construction.

\noindent \textbf{Pointwise selection fragments coding items.}
We sample 30K code responses from OpenCodeInstruct~\citep{ahmad2025opencodeinstruct} and use a frozen Qwen2.5-Coder-1.5B-Instruct~\citep{hui2024qwen2} to compute token-level CE scores. At each budget, the top-scoring code positions $A$ are mapped to their enclosing items. We measure the fraction of touched items that are only partially covered and the closure expansion $R=|\operatorname{Cl}(A)|/|A|$, where $\operatorname{Cl}(A)=\bigcup_{u:M(u)\cap A\neq\emptyset}M(u)$. Figure~\ref{fig:motivating_analysis} shows that, at the top-10\% budget, 92.3\% of touched items are incomplete and materializing their local spans requires $7.02\times$ as many positions.

\noindent \textbf{From materialized-item loss to local parser--test failure.}
Code provides two mechanically checkable outcomes: parser acceptance and test-suite success~\citep{Aho:72,humaneval}. We analyze an extracted executable code region and identify each model-token sequence with its decoded source string. Let $\operatorname{Pass}_{\mathcal{T}}(z)\in\{0,1\}$ equal one exactly when $z$ executes and passes every test in $\mathcal{T}$, and zero for a syntax, runtime, or test failure. Let $y=(y_1,\ldots,y_T)\in\operatorname{Lang}(G)$ be a reference program with $\operatorname{Pass}_{\mathcal{T}}(y)=1$, and define the teacher-forced loss $\ell_t(\theta)=-\log p_\theta(y_t\mid x,y_{<t})$. For $M(u)=[s,e]$, greedily decode exactly $|M(u)|$ tokens from the reference prefix $y_{<s}$, treating an emitted end-of-sequence token as a mismatch. Denote the continuation by $\hat y_{M(u)}$, splice it into the reference as $\tilde y^{(u)}=y_{<s}\circ\hat y_{M(u)}\circ y_{>e}$, and define
\[
F_{G,\mathcal{T}}(\tilde y^{(u)})
=\mathbf{1}\!\left[
\tilde y^{(u)}\notin\operatorname{Lang}(G)
\ \lor\ 
\operatorname{Pass}_{\mathcal{T}}(\tilde y^{(u)})=0
\right].
\]
\begin{proposition}[Reference-conditioned local parser--test failure bound]
\label{prop:local_failure}
For any model parameters $\theta$ and any parser-guided coding-item span $M(u)$,
\begin{equation}
F_{G,\mathcal{T}}(\tilde y^{(u)})
\leq
\mathbf{1}[\hat y_{M(u)}\neq y_{M(u)}]
\leq
\min\!\left\{
1,
\frac{\sum_{t\in M(u)}\ell_t(\theta)}{\log 2}
\right\}.
\label{eq:local_failure_bound}
\end{equation}
\end{proposition}

The first inequality holds because exact item recovery restores the valid, test-passing reference program. For the second, the first greedy mismatch occurs under the same prefix as teacher forcing; the selected incorrect token has probability at least that of the reference token, so the reference-token probability is at most $1/2$ and its loss is at least $\log 2$. Appendix~\ref{app:proofs} gives the complete proof.

Let $S=\{t:m_t=1\}$ be the supervised positions. If $M(u)\subseteq S$, every loss term in the bound appears in the numerator of the masked objective; if $S\cap M(u)\subsetneq M(u)$, at least one term is omitted. For example, omitting \texttt{*} from \texttt{total = price * qty} leaves an operator loss outside the objective even though replacing it can preserve parsing but change the test outcome. This provides a local diagnostic motivation for allocating code supervision at item rather than isolated-token granularity.

\noindent \textbf{Matched-budget mini-SFT.}
On the same 30K subset, we fine-tune Qwen2.5-Coder-1.5B-Instruct for 3{,}000 steps and compare isolated top-CE supervision with materialized-item supervision. The item arm maps top-CE anchors to their $M(u)$ spans; if their union contains $K$ distinct code positions, the pointwise arm supervises exactly the top-$K$ CE positions. The two arms therefore use the same number of supervised code tokens. Across HumanEval, HumanEval+, MBPP, and MBPP+, materialized-item supervision improves the average by 11.70 points (Figure~\ref{fig:motivating_analysis}, right), supporting item-level allocation under a controlled budget.

\section{Method}

\begin{figure}[t]
    \vspace{-30pt}
    \centering
    \includegraphics[width=\linewidth]{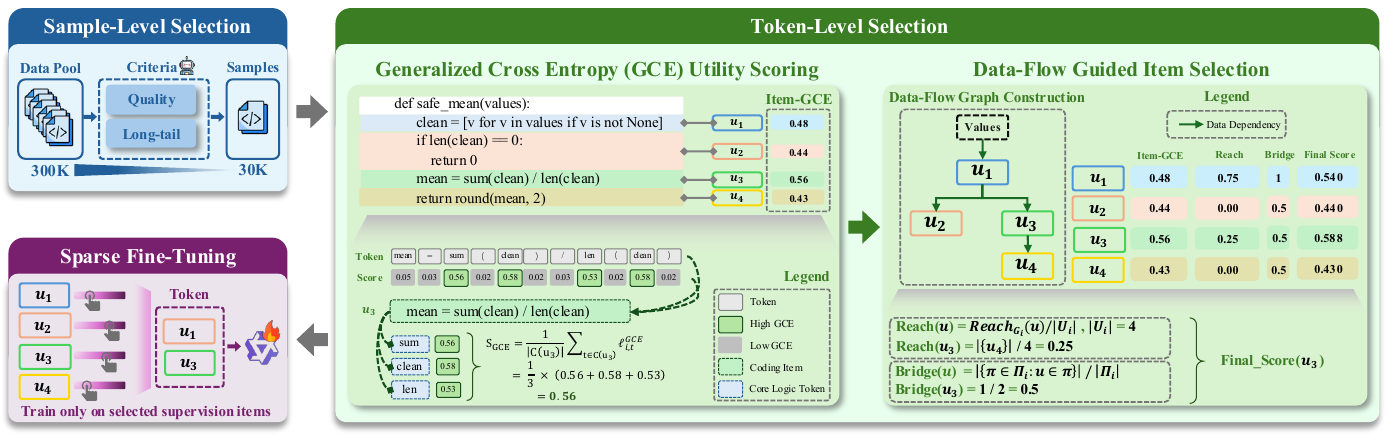}
    \caption{Overview of \method, which combines sample-level selection, GCE-based coding item scoring, data-flow guided item selection, and sparse fine-tuning for efficient code LLM supervised fine-tuning.}
    \label{fig:method_pipeline}
\end{figure}

In this section, we propose \method, a structure-aware sparse supervision method for supervised fine-tuning of code LLMs. As illustrated in Figure~\ref{fig:method_pipeline}, \method consists of four main components: sample-level selection, GCE-based coding item scoring, data-flow-guided item selection, and sparse fine-tuning.

\subsection{Sample-level Selection}

We first select a compact training subset from the large code instruction pool. Each instruction-response pair is scored by Qwen2.5-Coder-32B-Instruct \citep{hui2024qwen2} for instruction following and code correctness, and by a lightweight long-tail score based on structural/task features such as control flow, API usage, and code complexity. We rank sample $i$ by 
$s_i^{\mathrm{sample}}=\alpha s_i^{\mathrm{qual}}+(1-\alpha)s_i^{\mathrm{tail}}$,
Here, $s_i^{\mathrm{qual}}$ is a normalized quality score derived from LLM-judge ratings, with a small auxiliary calibration term from available unit test scores. The long-tail score $s_i^{\mathrm{tail}}$ measures the rarity of structural and task-level features, including API usage, control-flow patterns, and coarse code-complexity buckets. More selection details are provided in Appendix \ref{app:sample_selection}.
  After deduplication, the top 30K samples form $\mathcal{S}$ for coding-item-level supervision selection.

\subsection{GCE-Based Coding Item Scoring}
\label{sec:gce_scoring}
After sample selection, \method selects where to apply supervision within each response. For natural-language regions, we directly apply token-level GCE scoring and keep the top-$\rho_{\text{NL}}$ fraction of tokens. 
For code regions, we first construct coding items through a lightweight parser-guided local-closure procedure, as detailed in Algorithm~\ref{alg:code_unit_construction}. 
Specifically, we extract code regions from each response and parse them with Tree-sitter\footnote{\url{https://tree-sitter.github.io/tree-sitter/}}, then align syntax spans back to model-token indices using character offsets. 
Each token is labeled as a core logic token, a protected syntax token, or an other token. Core logic tokens, such as identifiers, literals, arithmetic/comparison operators, and API names, form the core set $C(u)$ of a coding item. 
Protected syntax tokens, such as brackets, commas, colons, delimiters, and statement boundaries, are retained when materializing $M(u)$, while other tokens act as separators. 
During the left-to-right scan, we close the current item when reaching a top-level statement boundary, crossing an unrelated separator, or reaching the end of the code region. For a terminated scan segment $[a,b]$, let $Q=C_{\mathrm{cur}}\cup\{j\in[a,b]:z_j=\mathrm{P}\}$. The \textsc{LocalClosure} rule is the deterministic interval $M(u)=[\min Q,\max Q]$: an $\mathrm{O}$ separator is excluded from $[a,b]$, while a closing boundary labeled $\mathrm{P}$ is included. Let $B(u)=M(u)\setminus C(u)$ denote the retained boundary and protected positions, so $M(u)=C(u)\mathbin{\dot\cup}B(u)$. Following the motivation in Sec.~\ref{sec:motivation}, we estimate utility over $C(u)$ but apply supervision to the materialized span $M(u)$.
For a coding item $u$, we define its utility by averaging token-level GCE scores over its core logic tokens:
\begin{equation}
S_{\mathrm{GCE}}(u)
=
\frac{1}{|C(u)|}
\sum_{t\in C(u)}
\ell_{i,t}^{\mathrm{GCE}},
\end{equation}
where $\ell_{i,t}^{\mathrm{GCE}}=\frac{1-p_{i,t}^{q}}{q}$,
$p_{i,t}=p_{\theta_0}(y_{i,t}\mid x_i,y_{i,<t}),$ and $\theta_0$ is the frozen base model. Here, $q\in(0,1]$ controls the degree of loss tempering.

We use GCE rather than raw CE to temper extremely low-probability outliers, which in code often correspond to rare identifiers, unusual literals, or noisy fragments. The item score is computed on $C(u)$ to measure uncertainty over logic-bearing tokens, while supervision is applied to the materialized span $M(u)$.

\subsection{Data-Flow Guided Item Selection}\label{sec:data_flow}

GCE identifies coding items that are uncertain under the base model, but uncertainty alone does not necessarily reflect structural importance. 
For example, a rare variable name or literal may obtain a high GCE score while affecting no later computation, whereas an intermediate assignment with a moderate GCE score may be reused by several subsequent statements and eventually determine the returned value. 
Therefore, \method further incorporates data-flow structure to prioritize items that are both learnable and structurally consequential. Following the code-representation view that definition--use relations expose how values propagate toward later computations~\citep{guo2020graphcodebert}, the two signals below operationalize downstream reach and path connectivity as lightweight ranking cues. They enter the priority as a gated bonus with a small weight rather than as the primary utility criterion.

For each code response, we construct a lightweight data-flow graph 
$\mathcal{G}_i=(\mathcal{U}_i,\mathcal{E}_i)$, where each node is a coding item and an edge $(u,v)\in\mathcal{E}_i$ indicates that item $v$ uses a value defined or updated by item $u$. 
Based on this graph, we compute two normalized structural signals:

\begin{itemize}[leftmargin=*]
    \item \textbf{Reach} measures the downstream influence of an item. 
    It captures whether the value defined or updated by item $u$ affects many later computations. 
    Formally, we define 
    $\operatorname{reach}(u)=|\mathrm{Reach}_{\mathcal{G}_i}(u)|/|\mathcal{U}_i|$, 
    where $\mathrm{Reach}_{\mathcal{G}_i}(u)$ denotes the set of items reachable from $u$ in $\mathcal{G}_i$.

    \item \textbf{Bridge} measures dependency-path connectivity. 
    It captures whether item $u$ lies on important dependency chains that connect early definitions to terminal computations. 
    Let $\Pi_i$ denote the set of source-to-sink paths in $\mathcal{G}_i$, where source nodes introduce values and sink nodes correspond to return statements, printed outputs, or final assignments. 
    We define 
    $\operatorname{bridge}(u)=|\{\pi\in\Pi_i:u\in\pi\}|/|\Pi_i|$ when $\Pi_i\neq\emptyset$, and $\operatorname{bridge}(u)=0$ otherwise.
\end{itemize}

We combine GCE utility with these data-flow signals through a gated priority function:
{\small
\begin{equation}
\begin{aligned}
& P_{\textsc{\method}}(u)
=  S_{\mathrm{GCE}}(u)
\Bigl[
1 + g(u)\lambda
\bigl(
\alpha_r \operatorname{reach}(u) 
+
\alpha_b \operatorname{bridge}(u)
\bigr)
\Bigr].
\end{aligned}
\end{equation}
}
where $g(u)=\mathbf{1}\left[S_{\mathrm{GCE}}(u)\ge Q_{\eta}^{(i)}\right]$. 
Here, $Q_{\eta}^{(i)}$ is the within-response GCE threshold for selecting the top-$\eta$ fraction of coding items, $\lambda$ controls the overall strength of the data-flow bonus, and $\alpha_r$ and $\alpha_b$ control the relative contributions of reach and bridge. 
Because our data-flow graph is constructed by lightweight static analysis, its structural signals may be approximate. The gate prevents structurally central but low-utility items from being over-selected due to noisy data-flow estimates, ensuring that data-flow only reranks items that are already sufficiently informative under GCE.

Finally, we derive the code-side supervision mask from the priority scores. Let $\operatorname{Top}_{\rho_{\mathrm{code}}}(\mathcal{U}_i;P_{\textsc{\method}})$ return the highest-priority coding items whose materialized spans $M(u)$ fit within the code-token budget ratio $\rho_{\mathrm{code}}$. The selected code positions and the corresponding supervision mask are defined as:
\begin{equation}
\begin{aligned}
\mathcal{M}^{\mathrm{code}}_i
&=
\bigcup_{u\in 
\operatorname{Top}_{\rho_{\mathrm{code}}}
(\mathcal{U}_i;P_{\textsc{\method}})
}
M(u), 
m^{\mathrm{code}}_{i,t}
=
\mathbf{1}\left[t\in \mathcal{M}^{\mathrm{code}}_i\right].
\end{aligned}
\end{equation}
Here, $M(u)$ is the materialized local span of item $u$. 
Tokens with $m^{\mathrm{code}}_{i,t}=1$ contribute to the sparse next-token prediction loss, while the remaining code tokens are kept as autoregressive context but ignored in the loss computation.

\begin{table}[t]
    \centering
    \vspace{-40pt}
    \caption{
    Performance comparison of different baselines on various benchmarks.
    We highlight the best result in \textbf{boldface} and the second-best with \underline{underline}.
    }
    \resizebox{\linewidth}{!}{
    \begin{tabular}{l|ccccccc|c}
\toprule
\textbf{Method} 
& \textbf{HumanEval} 
& \textbf{HumanEval+} 
& \textbf{MBPP} 
& \textbf{MBPP+} 
& \textbf{BCB-hard} 
& \textbf{BCB-full} 
& \textbf{Eff. Tokens (\%)} 
& \textbf{Average} \\
\midrule

\multicolumn{9}{c}{\cellcolor{gray!10} \textbf{Base model: Qwen2.5-Coder-1.5B-Instruct}} \\
\midrule
\textsc{Base} 
& 71.3 & 67.1 & 69.6 & 59.8 & 4.1 & 24.8 & -- & 49.4 \\

\textsc{Full Tokens} 
& \underline{75.6} & \underline{69.5} & 73.3 & 62.4 & 5.4 & 29.8 & 100.0 & 52.7 \\

\textsc{Random Selection} 
& 73.8 & 67.7 & \textbf{75.7} & \textbf{64.6} & 5.4 & \underline{30.4} & \underline{6.9} & 52.9 \\

\textsc{DS$^2$} 
& 74.4 & \underline{69.5} & 73.5 & 62.7 & \underline{7.4} & \textbf{31.2} & 17.0 & \underline{53.1} \\

\textsc{Token Cleaning} 
& 73.2 & 66.5 & 65.6 & 56.4 & 5.4 & 22.5 & \textbf{5.1} & 48.3 \\

\textsc{CLAM} 
& 73.2 & 68.9 & 74.1 & \underline{63.8} & \textbf{8.1} & 29.7 & 13.2 & 53.0 \\

\midrule

\textsc{\method} 
& \textbf{77.4} & \textbf{73.2} & \underline{75.4} & 63.2 & 6.1 & 29.9 & \underline{6.9} & \textbf{54.2} \\

\toprule
\multicolumn{9}{c}{\cellcolor{gray!10} \textbf{Base model: Seed-Coder-8B-Instruct}} \\
\midrule

\textsc{Base} 
& 78.7 & 68.3 & 82.0 & 69.0 & 25.7 & \underline{51.3} & -- & 62.5 \\

\textsc{Full Tokens}
& \underline{80.5} & 75.6 & 82.8 & 70.8 & \underline{28.3} & \textbf{52.8} & 100.0 & \underline{65.1} \\

\textsc{Random Selection} 
& 79.9 & \underline{76.2} & \textbf{86.5} & \textbf{73.8} & 25.0 & 45.1 & \underline{6.9} & 64.4 \\

\textsc{DS$^2$} 
& \underline{80.5} & \textbf{76.8} & \underline{86.0} & 73.0 & 25.7 & 45.1 & 17.0 & 64.5 \\

\textsc{Token Cleaning} 
& 72.0 & 65.9 & 77.5 & 65.3 & 14.9 & 37.3 & \textbf{5.1} & 55.5 \\

\textsc{CLAM} 
& 79.3 & 73.8 & \textbf{86.5} & \underline{73.3} & 25.7 & 44.8 & 13.2 & 63.9 \\

\midrule

\textsc{\method} 
& \textbf{82.3} & \textbf{76.8} & 85.7 & 71.7 & \textbf{30.4} & 46.0 & \underline{6.9} & \textbf{65.5} \\

\toprule
\multicolumn{9}{c}{\cellcolor{gray!10} \textbf{Base model: OpenCoder-8B-Base}} \\
\midrule

\textsc{Base} 
& 67.1 & 59.8 & 79.6 & 68.3 & 11.5 & \underline{40.9} & -- & 54.5 \\

\textsc{Full Tokens} 
& 64.0 & 60.9 & 78.0 & \textbf{68.7} & 12.8 & \textbf{42.8} & 100.0 & 54.5 \\

\textsc{Random Selection} 
& 70.7 & 62.8 & 79.9 & 68.3 & 11.5 & 40.7 & \underline{6.9} & 55.7 \\

\textsc{DS$^2$} 
& 71.3 & 63.4 & \textbf{80.7} & 67.5 & \textbf{15.5} & 38.7 & 17.0 & 56.2 \\

\textsc{Token Cleaning} 
& 71.3 & \underline{65.2} & 78.8 & 67.7 & 12.2 & 37.7 & \textbf{5.1} & 55.5 \\

\textsc{CLAM} 
& \underline{73.8} & \textbf{67.1} & 78.6 & \underline{68.5} & \underline{14.2} & 39.7 & 13.2 & \underline{57.0} \\

\midrule

\textsc{\method} 
& \textbf{76.2} & \textbf{67.1} & \underline{80.2} & 67.5 & 13.5 & 40.3 & \underline{6.9} & \textbf{57.5} \\

\bottomrule
\end{tabular}
    }
    \label{tab:main_results}
\end{table}

\subsection{Sparse Fine-Tuning}

Given the selected subset $\mathcal{S}$, we instantiate the sparse supervision mask defined in Sec.~3 by combining the code-side and natural-language-side masks. For code regions, we use $m^{\text{code}}_{i,t}$ from the data-flow-guided item selection in Sec.~5.3. For natural-language regions, we keep the top-$\rho_{\text{NL}}$ tokens according to GCE scores, denoted by $m^{\text{NL}}_{i,t}$. The final mask is

\begin{equation}
m_{i,t} = \mathbb{1}\left[m^{\text{code}}_{i,t}=1 \lor m^{\text{NL}}_{i,t}=1\right].
\end{equation}

We then optimize the masked next-token prediction objective in Eq.~(1), with unselected response tokens kept in the input sequence but ignored by the loss.

\begin{table}[t]
    \centering
    \caption{We remove key components of \method to evaluate their contributions. The best result is shown in \textbf{bold} and the second-best result is \underline{underlined}.}
    \resizebox{0.9\linewidth}{!}{
    \begin{tabular}{l|cccccc|c}
\toprule
\textbf{Method} 
& \textbf{HumanEval} 
& \textbf{HumanEval+} 
& \textbf{MBPP} 
& \textbf{MBPP+} 
& \textbf{BCB-hard} 
& \textbf{BCB-full} 
& \textbf{average} \\
\midrule
\multicolumn{8}{c}{\cellcolor{gray!10} \textbf{Base model: Qwen2.5-Coder-1.5B-Instruct}} \\
\midrule
\textsc{\method}  
& \textbf{77.4} 
& \textbf{73.2} 
& \textbf{75.4} 
& \underline{63.2} 
& \underline{6.1} 
& \underline{29.9} 
& \textbf{54.2} \\

\textsc{Token-off} 
& 73.2 
& 69.5 
& 73.3 
& \underline{63.2} 
& 5.4 
& \textbf{30.1} 
& 52.5 \\

\textsc{Sample-off} 
& \underline{76.8} 
& \underline{70.1} 
& 72.2 
& 62.4 
& 4.1 
& 29.2 
& 52.5 \\

\textsc{GCE-only} 
& \textbf{77.4} 
& \textbf{73.2} 
& \underline{74.6} 
& 63.0 
& 4.1
& 29.0 
& \underline{53.5} \\

\textsc{DataFlow-only} 
& 75.6 
& \underline{70.1} 
& 73.3 
& \textbf{63.5} 
& 2.7 
& 28.4 
& 52.3 \\
\bottomrule
    \end{tabular}
    }
    \label{tab:qwen25-coder-15b-ablation}
\end{table}

\section{Experiments}
\subsection{Experiments Setup}
\textbf{Dataset.} We construct our dataset by randomly sampling 300K instruction-tuning examples from OpenCodeInstruct~\citep{ahmad2025opencodeinstruct}. All training subsets and baseline variants are derived from this same pool to ensure a controlled comparison across data selection strategies. In the sample-level selection stage, we rank this pool and retain the top 30K examples as the curated training subset. 

\noindent \textbf{Base Models.} We evaluate our method on a diverse set of open-source code and general instruction-tuned LLMs, including Qwen2.5-Coder-1.5B-Instruct, Qwen2.5-Coder-3B-Instruct, Qwen2.5-Coder-7B-Instruct~\citep{hui2024qwen2}, Seed-Coder-8B-Instruct~\citep{seed2025seed}, and OpenCoder-8B-Base~\citep{huang2025opencoder}. These models are fine-tuned using samples selected from our data pool under the corresponding experimental settings.

\noindent \textbf{Baselines.} We compare our method against several representative data selection and token selection baselines. 1) Full Tokens fine-tunes the model on all 300K examples from the source pool, with supervision applied to every response token. 2) Random samples training examples or response tokens uniformly at random under the same data or token budget. 3) DS2~\citep{ds2} selects a curated subset using score-based data selection that favors high-quality and diverse examples. 4) Token Cleaning~\citep{pang2025tokencleaning} removes low-utility tokens according to a fixed-model token-level filtering criterion before fine-tuning. 5) CLAM~\citep{csam} performs clustering- and diversity-based sample selection to construct a representative training subset.
All baselines are trained with the same data pool, the same one-epoch budget, and the same optimizer configuration, and differ only in which samples or response tokens contribute to the loss.

\noindent \textbf{Evaluation.} We evaluate the fine-tuned models on six code generation benchmarks, including HumanEval, HumanEval+~\citep{humaneval}, MBPP, MBPP+~\citep{mbpp}, BigCodeBench-Hard, and BigCodeBench-Full~\citep{bcb}. These benchmarks provide a broad assessment of code generation performance, covering functional correctness, instruction following, and robustness across programming tasks of varying difficulty. We report pass@1 for each benchmark and use the average score over the six tasks as the main
 metric. Unless otherwise specified, we use the default settings in Table~\ref{tab:selection_hyperparams} for all experiments. More evaluation and training details are provided in Appendix~\ref{app:eval_setting}.

\subsection{Main Results}

\textbf{\method achieves the best performance--token trade-off.}
\method achieves the best average performance across all three models using only 6.9\% effective supervised tokens.
On Qwen2.5-Coder-1.5B-Instruct, Seed-Coder-8B-Instruct, and OpenCoder-8B-Base, \method obtains average scores of 54.2, 65.5, and 57.5, outperforming Full Tokens by 1.5, 0.4, and 3.0 points, respectively.
Under the same supervision budget, \method also consistently outperforms Random Selection, indicating that the gains arise from selecting more informative supervision targets rather than sparsity alone.
It further surpasses DS$^2$ and CLAM while using substantially fewer supervised tokens, demonstrating a better performance--efficiency trade-off than dense full-token supervision and existing selection baselines.

\textbf{Parser-guided item allocation benefits token selection in code.}
Although Token Cleaning is designed as a general token-level data selection method, its performance is unstable in the code domain.
It underperforms the corresponding base model on two of the three model settings, with average scores dropping from 49.4 to 48.3 on Qwen2.5-Coder-1.5B-Instruct and from 62.5 to 55.5 on Seed-Coder-8B-Instruct.
This observation suggests that directly selecting isolated tokens may be less suitable for code, where the usefulness of a token is often coupled with the local syntactic and semantic structure in which it appears.
In contrast, \method allocates supervision at the level of parser-guided code items and preserves structurally coherent supervision regions, resulting in substantially more stable improvements across different models and benchmarks.

\textbf{\method is effective on both base and instruction-tuned models.}
As shown in Table~\ref{tab:main_results}, \method achieves the highest average performance across both instruction-tuned models and the OpenCoder-8B base model.
This suggests that the effectiveness of our sparse supervision strategy is not restricted to a particular model initialization: it can further improve models that have already undergone instruction tuning while also providing substantial gains for a base model.
Additional results on Qwen2.5-Coder-3B-Instruct and Qwen2.5-Coder-7B-Instruct are reported in Table~\ref{tab:add_results_main}, further demonstrating the generality of \method across model scales.

\begin{figure}[t]
    \vspace{-30pt}
    \centering
    \begin{minipage}[t]{0.45\linewidth}
        \centering
        \includegraphics[width=\linewidth]{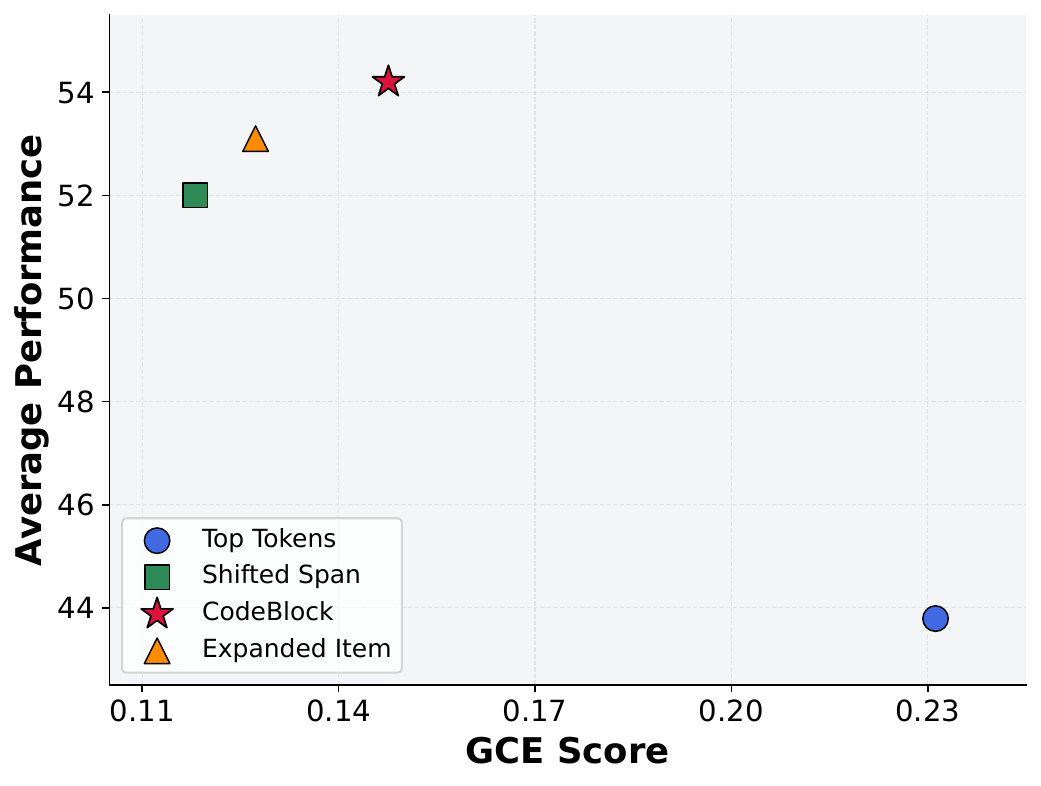}
        \caption{
        Performance under different code supervision granularities as a function of mean GCE score.
        }
        \label{fig:granularity_ablation}
    \end{minipage}
    \hfill
    \begin{minipage}[t]{0.48\linewidth}
        \centering
        \includegraphics[width=\linewidth]{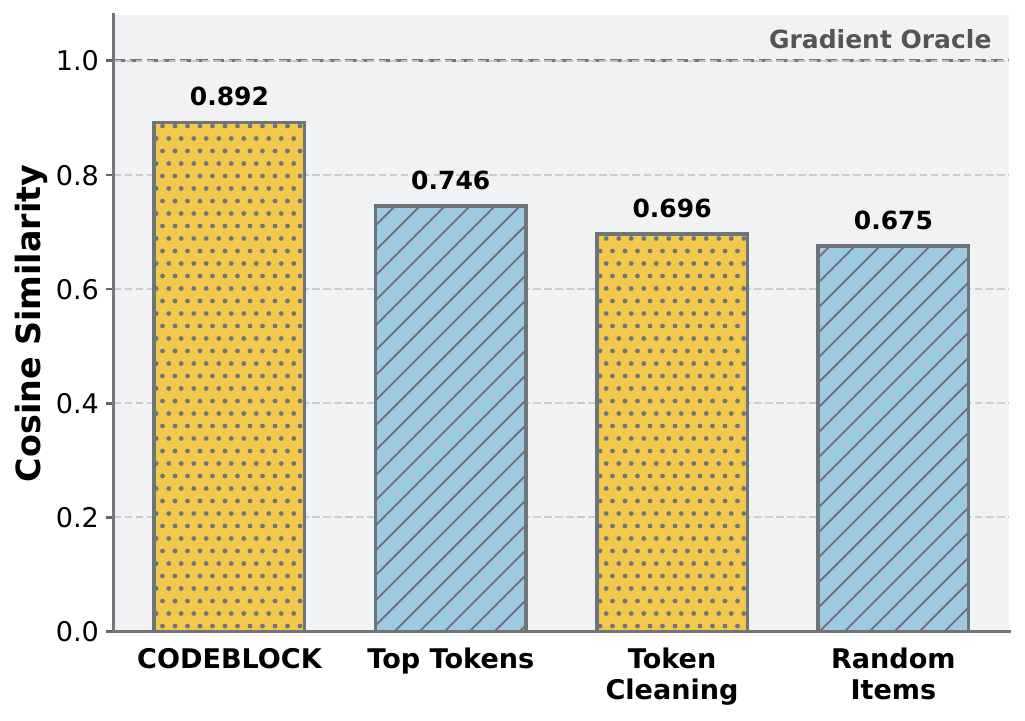}
        \caption{
        Cosine similarity between each method's aggregate gradient and the per-item Gradient Oracle.
        }
        \label{fig:gradient_oracle_ablation}
    \end{minipage}
    \vspace{-10pt}
\end{figure}

\begin{figure}[t]
    \vspace{-28pt}
    \centering
    \begin{minipage}[t]{0.38\linewidth}
        \centering
        \includegraphics[width=\linewidth]{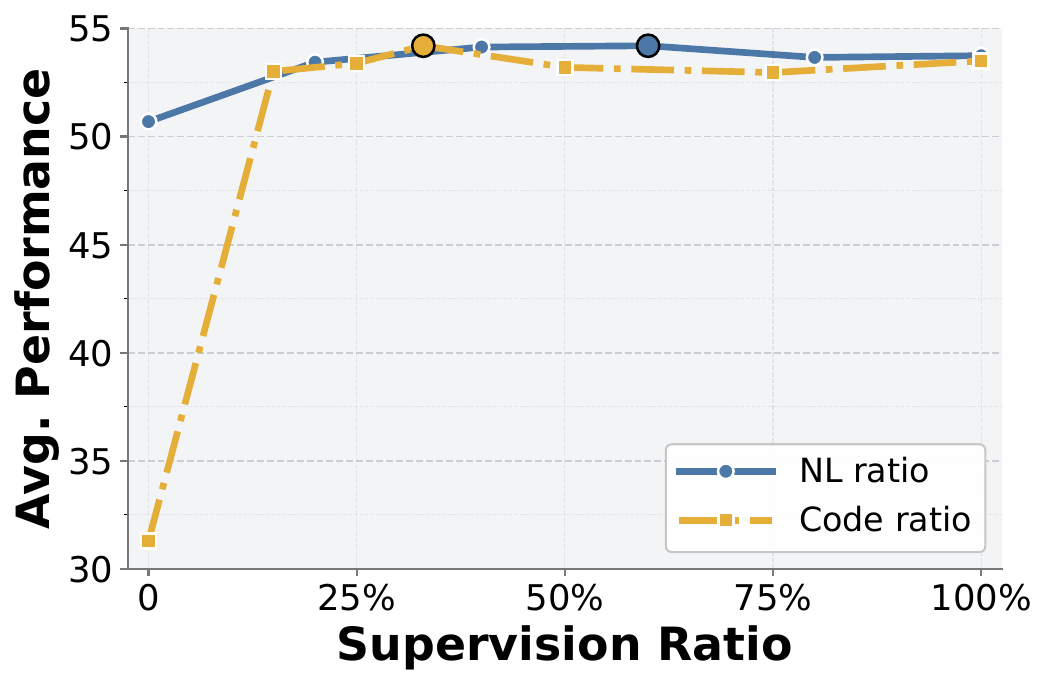}
        \caption{Sensitivity analysis of NL and code supervision ratios. }
        \label{fig:ratio_sensitivity}
    \end{minipage}
    \hfill
    \begin{minipage}[t]{0.60\linewidth}
        \centering
        \includegraphics[width=\linewidth]{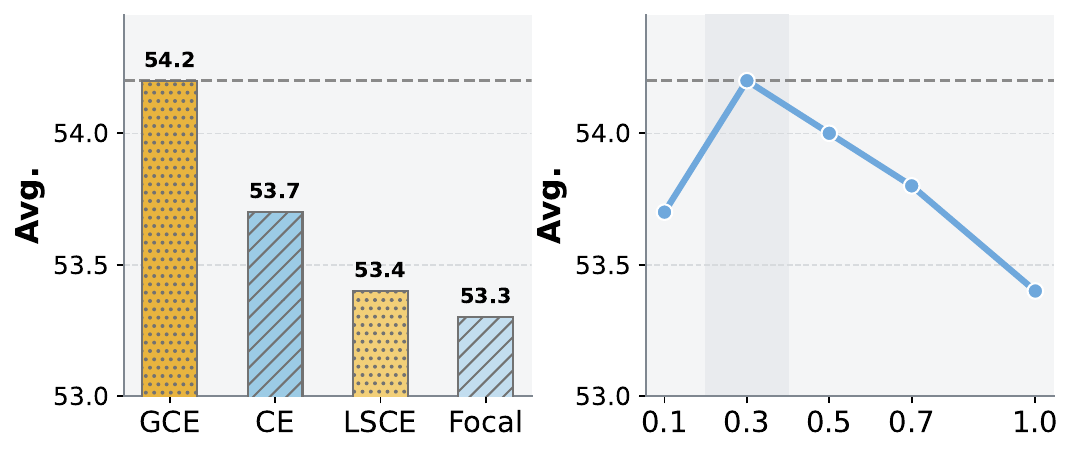}
        \caption{
Sensitivity analysis of token scoring sources (left) and the gate ratio $\eta$ (right).
}
        \label{fig:scoring_ablation}
    \end{minipage}
    \vspace{-10pt}
\end{figure}

\subsection{Ablation Study}
\textbf{Component Ablation Study.}
We conduct component ablations on Qwen2.5-Coder-1.5B-Instruct under the same training and evaluation setting as the main experiment. Token-off removes token-level selection, sample-off uses randomly selected samples, while GCE-only and dataflow-only retain only the corresponding scoring components.
As shown in Table~\ref{tab:qwen25-coder-15b-ablation}, removing any component reduces the average performance, with \method achieving the best overall result. GCE-only performs better than dataflow-only, suggesting that GCE provides the primary utility signal while data-flow information offers complementary structural guidance.

\textbf{What Is the Right Granularity for Code Supervision?}
In our main experiments, we use parser-aligned code items as the basic units of supervision. To examine whether this granularity is appropriate, we compare different supervision granularities under the same supervision budget.
Top Tokens supervises isolated tokens with the highest GCE scores; Shifted Span groups them into contiguous, length-matched spans while shifting the spans away from the original item boundaries; \method preserves complete parser-aligned code items; and Expanded Item further includes their directly neighboring items.
As shown in Figure~\ref{fig:granularity_ablation}, Top Tokens achieves the highest token-level GCE scores but performs substantially worse than the structurally coherent variants. \method achieves the best overall performance, while both fragmented supervision and unnecessary expansion are less effective. These results suggest that effective code supervision depends not only on token-level utility, but also on preserving an appropriate level of local structural coherence.

\textbf{Approximating the Gradient Oracle.}
We further examine whether \method can recover the selection behavior of an expensive per-item Gradient Oracle without explicitly computing gradients for every candidate item. Specifically, we construct the oracle over 23,346 coding items from 1,024 fixed candidate examples using a held-out APPS gradient, and compare all methods under the same code-only supervision space and 33\% per-example budget across 128 paired eight-example shards. We measure the cosine similarity between each method's aggregate gradient and the oracle-selected gradient. As shown in Figure~\ref{fig:gradient_oracle_ablation}, \method achieves substantially higher cosine similarity than both token-level baselines and random selection. These results suggest that structured coding-item selection provides an effective low-cost proxy for gradient-based oracle selection, capturing substantially more of the oracle-induced optimization direction than isolated token-level selection.

\textbf{Sensitivity Analysis.}
We analyze the sensitivity of \method{} to supervision ratios, token scoring, and gating, as shown in Figures~\ref{fig:ratio_sensitivity} and~\ref{fig:scoring_ablation}.
For supervision ratios, performance improves as more informative NL and code tokens are retained, peaking at the default $\rho_{\mathrm{NL}}=0.6$ and $\rho_{\mathrm{code}}=0.33$; further increasing either ratio provides no additional benefit, suggesting that excessive supervision introduces redundant tokens.
For token scoring, GCE performs best, outperforming CE, LSCE, and focal loss.
Similarly, a moderate gate ratio is preferable, with $\eta=30\%$ achieving the best performance, while both smaller and larger values degrade results.
Overall, these results support using selective supervision, GCE-based scoring, and moderate data-flow gating.
\section{Conclusion}

In this paper, we proposed \method, a structure-aware sparse supervision framework for code LLM fine-tuning. \method selects materialized parser-guided coding items rather than isolated high-scoring tokens, and prioritizes them with GCE utility and data-flow signals. Experiments across multiple code-generation benchmarks show that \method achieves competitive or better performance than full-token SFT and strong selection baselines while using far fewer supervised response tokens. These results support preserving parser-guided local spans and data-flow context in sparse code supervision, rather than treating tokens as independent learning targets. By keeping full responses as context and applying loss only to informative, dependency-critical code fragments, \method decouples contextual exposure from gradient supervision and offers a practical route toward efficient code model adaptation.

\section*{AI use statement}
In this work, we used generative AI tools for language polishing and for assisting with code drafting and debugging. All AI-assisted content was reviewed and revised by the authors, and generated code was verified through manual inspection and experimental testing. We take responsibility for the final content of this work, including text, claims or artifacts produced with the aid of generative AI.

\section*{Reproducibility statement}

We will release the source code and experimental implementation upon acceptance of the paper to facilitate reproducibility and future research.

\bibliographystyle{iclr2027_conference}
\bibliography{custom}
\newpage
\appendix

\section{Proofs for the Motivating Analysis}
\label{app:proofs}

Throughout, let $y\in\operatorname{Lang}(G)$ be a reference program with $\operatorname{Pass}_{\mathcal{T}}(y)=1$, let $M(u)=[s,e]$ be a parser-guided coding-item span, and let $\hat y_{M(u)}$ be the $|M(u)|$-token greedy continuation decoded from $y_{<s}$.

\begin{proof}[Proof of Proposition~\ref{prop:local_failure}]
If $\hat y_{M(u)}=y_{M(u)}$, then $\tilde y^{(u)}=y$, so $F_{G,\mathcal{T}}(\tilde y^{(u)})=0$. Since $F_{G,\mathcal{T}}$ is binary, this proves the first inequality.

For the second inequality, the result is immediate when $\hat y_{M(u)}=y_{M(u)}$. Otherwise, let
\[
t^\star=\min\{t\in M(u):\hat y_t\neq y_t\}
\]
be the first mismatched position. Local decoding starts from $y_{<s}$, and all locally generated tokens before $t^\star$ match their reference tokens; hence the decoder prefix at $t^\star$ is exactly $y_{<t^\star}$. Because greedy decoding chooses an incorrect token $\hat y_{t^\star}$,
\[
p_\theta(\hat y_{t^\star}\mid x,y_{<t^\star})
\geq
p_\theta(y_{t^\star}\mid x,y_{<t^\star}).
\]
The probabilities of these two distinct tokens sum to at most one, so $p_\theta(y_{t^\star}\mid x,y_{<t^\star})\leq 1/2$ and therefore $\ell_{t^\star}(\theta)\geq\log 2$. It follows that
\[
\mathbf{1}[\hat y_{M(u)}\neq y_{M(u)}]
\leq
\frac{\sum_{t\in M(u)}\ell_t(\theta)}{\log 2}.
\]
Combining this inequality with the trivial upper bound of one proves Eq.~\ref{eq:local_failure_bound}.
\end{proof}


\section{Details Algorithm of Coding Item Construction}
\label{sec:algorthim}
\vspace{-2pt}
\begin{algorithm}[h]
\small
\caption{\method Coding-Item Construction with Local Closure}
\label{alg:code_unit_construction}
\hrule
\vspace{1pt}
\begin{algorithmic}[1]
\Require Response $y_i$, response-token indices $\mathcal{I}_i$ with character offsets, code regions $\mathcal{F}_i$, protected syntax set $\Omega_{\mathrm{P}}$
\Ensure Coding items $\mathcal{U}_i=\{u_{i,k}\}_{k=1}^{K_i}$, where each item $u=(C(u),M(u))$

\State $\mathcal{U}_i \gets \emptyset$
\For{each code region $f\in\mathcal{F}_i$}
    \State $z \gets \textsc{LabelCodeTokens}(f,\mathcal{I}_i,\Omega_{\mathrm{P}})$
    \Comment{$z_r\in\{\mathrm{C},\mathrm{P},\mathrm{O}\}$}
    \State $C_{\mathrm{cur}}\gets \emptyset$, $d\gets 0$, $a\gets$ first token index in $f$
    \For{each token index $r$ in $f$ from left to right}
        \If{$z_r=\mathrm{C}$}
            \State $C_{\mathrm{cur}}\gets C_{\mathrm{cur}}\cup\{r\}$
        \ElsIf{$C_{\mathrm{cur}}\neq\emptyset$}
            \If{\textsc{Boundary}$(t_r,d)$ or $z_r=\mathrm{O}$}
                \State $b\gets r$ if $z_r=\mathrm{P}$, else $b\gets r-1$
                \State $M\gets \textsc{LocalClosure}(C_{\mathrm{cur}},[a,b],z)$
                \State $\mathcal{U}_i\gets \mathcal{U}_i\cup\{(C_{\mathrm{cur}},M)\}$
                \State $C_{\mathrm{cur}}\gets\emptyset$
            \EndIf
        \EndIf
        \If{\textsc{Boundary}$(t_r,d)$ or $z_r=\mathrm{O}$}
            \State $a\gets r+1$
        \EndIf
        \State $d\gets \textsc{UpdateDepth}(t_r,d)$
    \EndFor
    \If{$C_{\mathrm{cur}}\neq\emptyset$}
        \State $b\gets$ last token index in $f$
        \State $M\gets \textsc{LocalClosure}(C_{\mathrm{cur}},[a,b],z)$
        \State $\mathcal{U}_i\gets \mathcal{U}_i\cup\{(C_{\mathrm{cur}},M)\}$
    \EndIf
\EndFor
\State \Return $\mathcal{U}_i$
\hrule
\end{algorithmic}
\end{algorithm}

\section{Experimental Details}
\label{app:ex_detail}
\subsection{Sample-Level Selection Settings}
\label{app:sample_selection}

We perform sample-level selection from the 300K OpenCodeInstruct (OCI) source pool before applying item-level sparse supervision. 
Instead of re-querying an external judge model, we reuse the LLM-judge scores provided by OCI and follow its scoring protocol. 
Each sample is evaluated along three dimensions: requirement conformance, logical correctness, and edge-case consideration. 
We compute the mean judge score over these dimensions and combine it with a lightweight test-based auxiliary score:
\begin{equation}
s^{\mathrm{raw}}_{\mathrm{qual}} 
= 
s_{\mathrm{judge}} + 0.2 \cdot s_{\mathrm{test}},
\end{equation}
where \(s_{\mathrm{judge}}\) denotes the average LLM-judge score and \(s_{\mathrm{test}}\) denotes the average available test score. 
The test score is used only as a small auxiliary signal for tie-breaking and calibration, rather than as the dominant selection criterion. 
We then apply robust min--max normalization to obtain the normalized quality score \(s^{\mathrm{norm}}_{\mathrm{qual}} \in [0,1]\).

To avoid selecting only common and easy high-quality examples, we further compute a lightweight long-tail score. 
The long-tail score combines tag-level rarity and bucket-level rarity:
\begin{equation}
s_{\mathrm{tail}}
=
0.7 \cdot s_{\mathrm{tag}}
+
0.3 \cdot s_{\mathrm{bucket}}.
\end{equation}
Here, \(s_{\mathrm{tag}}\) captures the presence of relatively rare program structures or API-related tags, such as imports, classes, multiple functions, recursion, exception handling, graph-related operations, date/time operations, and asynchronous programming. 
The bucket score \(s_{\mathrm{bucket}}\) measures rarity with respect to prompt length, code length, and coarse complexity buckets.

The final sample-level ranking score is computed as:
\begin{equation}
s_{\mathrm{sample}}
=
0.7 \cdot s^{\mathrm{norm}}_{\mathrm{qual}}
+
0.3 \cdot s_{\mathrm{tail}}.
\end{equation}

\subsection{Token and Item Selection Settings}
\label{app:token_item_selection_settings}

Table~\ref{tab:selection_hyperparams} summarizes the default hyperparameters used in the token- and item-level selection stage of \method. 
Unless otherwise specified, all experiments use the same configuration.

\begin{table}[t]
\caption{Default hyperparameters for token- and item-level supervision selection in \method.}
\centering
\small
\setlength{\tabcolsep}{5pt}
\renewcommand{\arraystretch}{1.12}
\begin{tabular}{p{0.20\linewidth} p{0.52\linewidth} p{0.18\linewidth}}
\hline
\textbf{Hyperparameter} & \textbf{Description} & \textbf{Default} \\
\hline
$q$ 
& Smoothing parameter in the GCE score. 
& $0.7$ \\

$\rho_{\mathrm{NL}}$ 
& Retention ratio for natural-language tokens. 
& $0.6$ \\

$\rho_{\mathrm{code}}$ 
& Code supervision budget ratio for selected coding items. 
& $0.33$ \\

$\eta$ 
& \method gate ratio; only the top-$\eta$ GCE-scored items receive data-flow bonuses. 
& $30\%$ \\

$\lambda$ 
& Overall strength of the data-flow bonus. 
& $0.10$ \\

$\alpha_r$ 
& Weight of the reach signal. 
& $1.00$ \\

$\alpha_b$ 
& Weight of the bridge signal. 
& $0.50$ \\
\hline
\end{tabular}
\label{tab:selection_hyperparams}
\end{table}

\subsection{Fine-Tuning Settings}
\label{app:finetuning_settings}

We report the main fine-tuning settings in Table~\ref{tab:finetuning_settings}.
All methods, including full-token SFT, \method, and every selection baseline,
are fine-tuned for exactly one epoch over the training set produced by the
corresponding selection strategy under an identical optimization setup.
All experiments use an effective batch size of 8 and are conducted on one
NVIDIA A800 GPU with bf16 precision.

For sparse-supervision variants, the input sequence is kept unchanged: unselected response tokens remain available as autoregressive context, but their labels are set to the ignore index and therefore do not contribute to the cross-entropy loss. 
Within the selected training examples, \method differs from full-token SFT only in the supervision mask, while using the same causal language modeling
objective.

\begin{table}[t]
\caption{Fine-tuning settings used for the main experiments.}
\centering
\small
\begin{tabular}{l|l}
\toprule
Setting & Value \\
\midrule
Training budget & 1 epoch for all methods \\
Learning rate schedule & LR = $3\times10^{-5}$, scheduler = cosine \\
Effective batch size & 8 = per-device batch size 2 $\times$ grad. accumulation 4 $\times$ GPUs 1 \\
Maximum sequence length & 2048 tokens \\
Optimization setup & AdamW, weight decay 0.0, warmup ratio 0.05, bf16 \\
\bottomrule
\end{tabular}
\label{tab:finetuning_settings}
\end{table}

\subsection{Evaluation Settings}
\label{app:eval_setting}
For HumanEval and MBPP, we use EvalPlus and report sanitized pass@1 on HumanEval, HumanEval+, MBPP, and MBPP+. For BigCodeBench, we evaluate instruction-tuned models on the instruct split and base models on the complete split, reporting pass@1 on the hard and full subsets. All reported pass@1 scores use greedy decoding with temperature 0.0 and one generated sample per task.

\section{Runtime Analysis}
\label{app:runtime_analysis}
\begin{table}[t]
    \caption{
    Observed wall-clock runtime in minutes on the Qwen2.5-Coder-1.5B-Instruct
    setting. We include method-specific preprocessing and SFT training, and
    exclude evaluation.
    }
    \centering
    \small
    \setlength{\tabcolsep}{5pt}
    \begin{tabular}{@{}lrrrr@{}}
    \toprule
    \textbf{Method} & \textbf{Preprocess} & \textbf{Train} & \textbf{Total} & \textbf{Total / Full} \\
    \midrule
    Full Tokens (300K) & 8.9 & 257.7 & 266.6 & 100.0\% \\
    Random Selection & 3.2 & 37.6 & 40.8 & 15.3\% \\
    \method & 32.1 & 36.8 & 68.9 & 25.8\% \\
    Token Cleaning & 32.6 & 12.1 & 44.7 & 16.8\% \\
    DS$^2$ & 29.0 & 38.6 & 67.6 & 25.4\% \\
    CLAM & 16.5 & 35.5 & 52.0 & 19.5\% \\
    \bottomrule
    \end{tabular}
    \label{tab:runtime_analysis}
    \end{table}
We additionally report observed wall-clock runtime on the Qwen2.5-Coder-1.5B-Instruct main setting. 
For each method, we count method-specific preprocessing and SFT training time, while excluding evaluation time. 
Shared dataset construction costs, such as LLM-judge scoring for constructing the common candidate pool, are not attributed to any individual method.

As shown in Table~\ref{tab:runtime_analysis}, Full Tokens requires 266.6 minutes in total, with most of the cost coming from training on the full 300K dataset. 
In contrast, \method requires 68.9 minutes, or 25.8\% of the Full Tokens runtime. 
Although \method introduces additional preprocessing for token scoring and structured item construction, its reduced training workload substantially lowers the overall runtime. 
Its total runtime is comparable to DS$^2$ and remains far below full-token training, while being higher than Random Selection, Token Cleaning, and CLAM. 
Overall, the structure-aware preprocessing introduces moderate overhead while preserving a substantial runtime advantage over dense full-data supervision.

\section{Additional Experiment Results}
\label{append:add_results}

\subsection{More Experimental Results}
\label{app:more_results}

We report additional results on Qwen2.5-Coder-3B-Instruct and Qwen2.5-Coder-7B-Instruct to further examine the generality of \method. 
All methods use the same data pool, training protocol, and evaluation setting as the main experiments. 
As shown in Table~\ref{tab:add_results_main}, \method consistently achieves strong average performance with only about 6.9\% supervised response tokens, further supporting the effectiveness of structure-aware sparse supervision across different model scales.

\begin{table}[t]
\caption{
    Additional comparison of different baselines on various benchmarks.
    We highlight the best result in \textbf{boldface} and the second-best with \underline{underline}.
    }
    \centering
    \resizebox{\linewidth}{!}{
    \begin{tabular}{l|ccccccc|c}
\toprule
\textbf{Method} 
& \textbf{HumanEval} 
& \textbf{HumanEval+} 
& \textbf{MBPP} 
& \textbf{MBPP+} 
& \textbf{BCB-hard} 
& \textbf{BCB-full} 
& \textbf{Eff. Tokens (\%)} 
& \textbf{Average} \\
\midrule

\multicolumn{9}{c}{\cellcolor{gray!10} \textbf{Base model: Qwen2.5-Coder-3B-Instruct}} \\
\midrule

\textsc{Base} 
& 84.2 & \textbf{79.3} & 73.8 & 62.7 & 13.5 & 34.8 & -- & 58.0 \\

\textsc{Full Tokens} 
& 82.3 & 75.0 & \underline{81.2} & \textbf{69.6} & 14.9 & 37.2 & 100.0 & 60.0 \\

\textsc{Random Selection} 
& 83.5 & 75.6 & 78.3 & 65.9 & \textbf{16.9} & \underline{42.3} 
& \underline{6.9} & 60.4 \\

\textsc{DS$^2$} 
& \textbf{86.0} & \underline{78.1} & 79.1 & 66.7 & \textbf{16.9} & \textbf{43.3} 
& 17.0 & \textbf{61.7} \\

\textsc{Token Cleaning} 
& 82.9 & 76.8 & 78.6 & 64.8 & 12.2 & 36.8 
& \textbf{5.1} & 58.7 \\

\textsc{CLAM} 
& \underline{84.8} & \underline{78.1} & \textbf{81.5} & 66.7 & 14.2 & 41.8 
& 13.2 & 61.2 \\

\midrule

\textsc{\method} 
& 83.5 & 77.4 & \textbf{81.5} & \underline{68.0} & \underline{16.2} & 42.1 
& \underline{6.9} & \underline{61.5} \\

\toprule
\multicolumn{9}{c}{\cellcolor{gray!10} \textbf{Base model: Qwen2.5-Coder-7B-Instruct}} \\
\midrule

\textsc{Base} 
& 87.2 & 83.1 & 83.5 & 71.6 & 19.4 & 39.7 
& -- & 64.1 \\

\textsc{Full Tokens} 
& 87.8 & 82.9 & 85.2 & 71.4 & 18.9 & 39.9 
& 100.0 & 64.4 \\

\textsc{Random Selection} 
& 87.2 & \underline{83.5} & \textbf{87.8} & \underline{75.1} & 18.9 & 41.3 
& \underline{6.9} & 65.6 \\

\textsc{DS$^2$} 
& 86.0 & 81.1 & \underline{87.3} & 72.5 & \textbf{24.3} & \underline{48.8} 
& 17.0 & \underline{66.7} \\

\textsc{Token Cleaning} 
& 81.7 & 76.8 & 84.9 & 73.3 & 18.2 & 40.3 
& \textbf{5.1} & 62.5 \\

\textsc{CLAM} 
& \underline{89.0} & \textbf{84.2} & 86.5 & \underline{75.1} & 12.8 & 39.4 
& 13.2 & 64.5 \\

\midrule

\textsc{\method} 
& \textbf{90.2} & \underline{83.5} & 86.8 & \textbf{75.4} & \underline{20.3} & \textbf{49.2} 
& \underline{6.9} & \textbf{67.6} \\

\bottomrule
\end{tabular}
    }
    \label{tab:add_results_main}
\end{table}

\subsection{More Ablation Study}
\label{app:more_ablation}
\begin{wrapfigure}[17]{r}{0.52\textwidth}
    \centering
    \includegraphics[width=\linewidth]{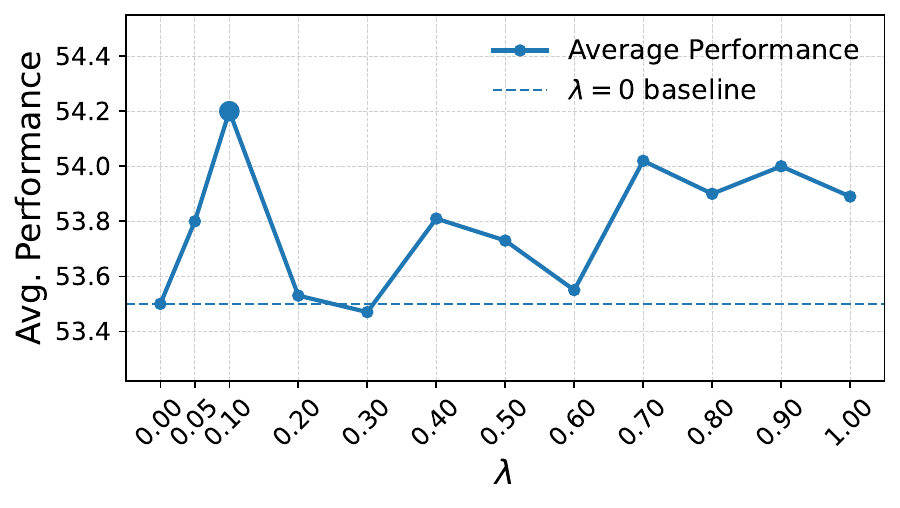}
    \caption{Sensitivity analysis of $\lambda$ on the average pass@1 score across six code-generation benchmarks.}
    \label{fig:lambda_ablation}
\end{wrapfigure}
We further study the effect of the data-flow bonus strength $\lambda$ in the \method priority function. 
This hyperparameter controls how strongly the reach and bridge signals influence the final ranking of candidate coding items. 

Figure~\ref{fig:lambda_ablation} reports the results under different values of $\lambda$, while keeping all other hyperparameters fixed. 
The results show that incorporating the data-flow bonus is generally beneficial, with $\lambda=0.10$ achieving the best average performance of $54.2$. 
Compared with the $\lambda=0$ baseline of $53.5$, this improvement suggests that reach and bridge signals provide complementary structural information beyond token-level GCE utility. 

However, the trend is not monotonic, and further increasing $\lambda$ does not consistently improve performance. 
While most non-zero settings outperform the $\lambda=0$ baseline, overly emphasizing the structural bonus can reduce the benefit. 
One possible reason is that the data-flow graph is constructed through static code analysis and therefore provides only an approximate characterization of dependency importance, particularly for incomplete snippets, library calls, and implicit dependencies. 
Thus, data-flow information is more effective as a lightweight reranking signal that complements GCE utility rather than dominating the original ranking. 
Overall, these results support the effectiveness of data-flow-aware reranking while highlighting the importance of properly balancing structural and token-level utility signals. 
We therefore set $\lambda=0.10$ as the default value in our experiments.

\end{document}